\documentclass[letterpaper, 10 pt, conference]{ieeeconf}  

\IEEEoverridecommandlockouts                              

\usepackage{amsmath,amsfonts}
\usepackage{amssymb}

\usepackage{algorithm}
\usepackage[noend]{algorithmic}
\usepackage{color}

\usepackage{array}
\usepackage{textcomp}
\usepackage{stfloats}

\usepackage{verbatim}
\usepackage{graphicx}

\usepackage{cite}
\usepackage[utf8]{inputenc} 
\usepackage[T1]{fontenc}    
\usepackage{hyperref}       
\usepackage{url}            
\usepackage{booktabs}       
\usepackage{amsfonts}       
\usepackage{nicefrac}       
\usepackage{microtype}      
\usepackage{lipsum}
\usepackage{fancyhdr}       
\usepackage{graphicx}       
\usepackage{wrapfig}
\usepackage{subfigure}
\usepackage{multirow}
\usepackage{threeparttable}

\usepackage{xcolor}
\definecolor{customgreen}{HTML}{8ED973}

\title{\LARGE \bf \textbf{ArtGS}:
3D Gaussian Splatting for Interactive Visual-Physical Modeling and Manipulation of Articulated Objects
}

\author{ Qiaojun Yu$^{1, 2*}$, Xibin Yuan$^{1*}$, Yu Jiang$^{1*}$, Junting Chen$^{3}$, \\ 
Dongzhe Zheng$^{4}$, Ce Hao$^{3}$, Yang You$^{5}$, Yixing Chen$^{1}$, Yao Mu$^{1}$,  Liu Liu$^{6}$ and Cewu Lu$^{1\dagger}$ 
\thanks{$^*$ indicates the equal contribution. $^\dagger$ Cewu Lu is the corresponding author, \texttt{lucewu@sjtu.edu.cn}. $^1$Qiaojun Yu,  Xibin Yuan, Yu Jiang, Yixing Chen, and Cewu Lu are with the Shanghai Jiao Tong University, China. $^2$Qiaojun Yu is with Shanghai AI Laboratory, China. $^3$Ce Hao and Junting Chen are with National University of Singapore, Singapore. $^4$Dongzhe Zheng is with Princeton University, USA. $^5$Yang You is with Stanford University, USA. $^6$Liu Liu is with Hefei University of Technology, China.}
\thanks{\textbf{Acknowledgment}: This work is supported by the National Natural Science Foundation of China under Grant (No.62302143), by the National Key R\&D Program of China (Grant No.2024YFB4707600), by the Shanghai Commitee of Science and Technology, China(Grant No.24511103200), by the National Key Research and Development Project of China (No.2022ZD0160102), Shanghai Artificial Intelligence Laboratory, XPLORER PRIZE grants.}
}

\begin{document}

\maketitle


\begin{abstract}
Articulated object manipulation remains a critical challenge in robotics due to the complex kinematic constraints and the limited physical reasoning of existing methods. In this work, we introduce ArtGS, a novel framework that extends 3D Gaussian Splatting (3DGS) by integrating visual-physical modeling for articulated object understanding and interaction. ArtGS begins with multi-view RGB-D reconstruction, followed by reasoning with a vision-language model (VLM) to extract semantic and structural information, particularly the articulated bones. Through dynamic, differentiable 3DGS-based rendering, ArtGS optimizes the parameters of the articulated bones, ensuring physically consistent motion constraints and enhancing the manipulation policy. By leveraging dynamic Gaussian splatting, cross-embodiment adaptability, and closed-loop optimization, ArtGS establishes a new framework for efficient, scalable, and generalizable articulated object modeling and manipulation. Experiments conducted in both simulation and real-world environments demonstrate that ArtGS significantly outperforms previous methods in joint estimation accuracy and manipulation success rates across a variety of articulated objects. Additional images and videos are available on the project website: 
\href{https://sites.google.com/view/artgs/home}{sites.google.com/view/artgs}.
\end{abstract}

\section{Introduction} \label{Sec: intro}

Previous approaches to robotic perception and manipulation, such as end-to-end reinforcement learning~\cite{wu2021vat} or imitation learning~\cite{mu2025robotwin, yu2025forcevla}, typically require intensive trial-and-error or large-scale demonstrations to generate manipulation trajectories. These methods often rely on extensive interaction data collected in complex environments, typically through simulation or teleoperation. Despite these efforts, these methods still struggle to deliver reliable performance. The generated manipulation trajectories often lack principled physical knowledge—especially regarding joint kinematics—resulting in constraint-violating motions and eventual manipulation failures during task execution. In contrast, 3D vision-based approaches explicitly construct articulation models using techniques such as part segmentation and joint parameter estimation. Although methods like GAMMA~\cite{yu2023gamma} and RPMArt~\cite{wang2024rpmart} have achieved notable progress in extracting spatial information, the inherently sparse and unordered nature of point clouds—along with their limited temporal consistency—continues to pose significant challenges.

\begin{figure}[t]
  \begin{center}
   \includegraphics[width=0.8\linewidth]{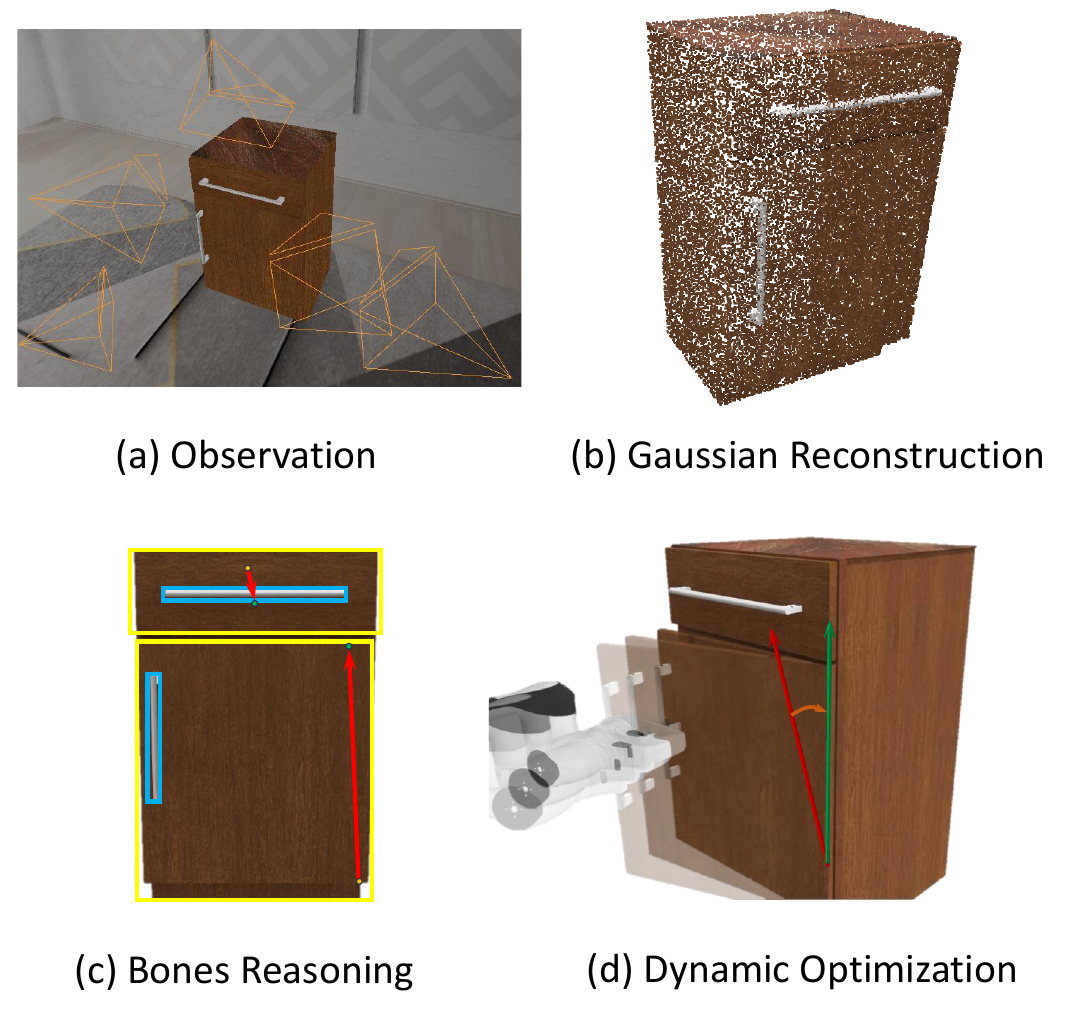}
  \end{center}
    \vspace{-10pt}
      \caption{\textbf{The overview of ArtGS.} (a) Multi-view observations of the articulated object. (b) Static reconstruction using 3DGS. (c) Inference of articulated bone parameters using the fine-tuned Vision-Language Model. (d) Optimization of parameters through robotic manipulation.}
\label{Fig: teaser}
\vspace{-15pt}
\end{figure}

To address these challenges, we propose a novel framework, \textbf{ArtGS}, which extends 3D Gaussian Splatting (3DGS)~\cite{kerbl20233d} by integrating visual reconstruction with physical modeling. It combines part-level 3DGS representations with articulation-aware physical constraints, leveraging dynamic motion information to optimize the corresponding physical parameters. By exploiting the particle-based nature of 3DGS, \textbf{ArtGS} maps Gaussian spheres to articulated bones, imposes explicit motion constraints through differentiable rendering and the inherent spatial flexibility of the spheres, and seamlessly integrates visual perception with physical modeling.

Specifically, \textbf{ArtGS} first performs high-fidelity 3D scene reconstruction from multi-view RGB-D images, followed by visual-language reasoning using fine-tuned vision-language models (VLMs) to facilitate articulated object understanding. This enables precise segmentation of articulation parts and the initial estimation of parameters corresponding to each articulated bone. Instead of relying on the \textbf{6D motion of individual Gaussian spheres}, ArtGS constructs articulated parts based on estimated bones and integrates them into a unified visual-physical model. The associated Gaussian spheres undergo coordinated \textbf{1D motion transformations, constrained by joint parameters}, forming the basis for subsequent interactive optimization. By leveraging the differentiable rendering capabilities of 3DGS and interactive manipulation with an embodied robot, \textbf{ArtGS} performs closed-loop optimization of bone parameters, integrating visual reconstruction with physical modeling to ensure adherence to physical constraints while preserving visual–physical consistency. Our main contributions are summarized as follows:
\begin{itemize}
    \item We propose \textbf{ArtGS}, a framework for interactive visual–physical modeling of articulated objects, which integrates static 3DGS reconstruction with a vision–language model (VLM), fine-tuned on a domain-specific dataset, to inject physical modeling into the visual perception process and enable interactive optimization of the visual–physical models—specifically, the articulated bone parameters—through manipulation.
    \item By transforming the robot into a high-fidelity 3D Gaussian Splatting (3DGS) digital twin through the use of Modified Denavit-Hartenberg (MDH) forward kinematics for dynamic reconstruction, \textbf{ArtGS} enables adaptable representations across various robotic embodiments. Leveraging temporal and spatial coherence in dynamic 3DGS and differentiable rendering, \textbf{ArtGS} effectively optimizes articulated bone parameters, mitigates occlusion effects, and enhances optimization through dynamic interaction sequences.
    \item We conduct extensive experiments in both simulated and real-world environments, showing that \textbf{ArtGS}—integrating visual-physical modeling and robotic interaction—substantially improves modeling accuracy and manipulation success rates for various articulated objects.
\end{itemize}

\section{Related Works} \label{Sec: related}

\subsection{Articulated Object Modeling}
Articulated object modeling has been extensively studied in the context of articulated object reconstruction~\cite{liCategoryLevelArticulatedObject2020, bozicNeuralDeformationGraphs2021, wu2025reartgs, zengMARSMultimodalActive2024, jiang2022ditto, Weng2024NeuralIR} and joint parameters estimation~\cite{jain2021screwnet, zeng2021visual, yu2024uniaff, chuCommanddrivenArticulatedObject2023}, with methods broadly categorized into explicit joint modeling methods with part-based reconstruction and implicit methods with neural representations. These paradigms address the challenges of understanding object geometry and articulation while enabling applications in robotics, animation, and interaction modeling. \textbf{Explicit methods} model articulated objects~\cite{yu2025generalizable, jiang2022ditto, geng2023gapartnet} in a more straightforward manner by leveraging geometric primitives, 3D part segmentation, and explicit joint kinematics.
PPD~\cite{kawanaUnsupervisedPoseawarePart2022} developed an unsupervised decomposition method for man-made objects using revolute and prismatic joint priors, enforcing kinematic consistency through pose-aware chamfer distances between canonical and articulated part spaces. MARS ~\cite{zengMARSMultimodalActive2024} introduced a multi-modal fusion module utilizing multi-scale RGB features to enhance point cloud features, coupled with active sensing for autonomous optimization of observation viewpoints for better segmentation and joint parameter estimation. Ditto~\cite{jiang2022ditto} and DigitalTwinArt~\cite{Weng2024NeuralIR} adopt explicit spatial motion representations to model previously unseen articulated objects from two RGB-D scans captured in different articulation states.
\textbf{Implicit methods} usually encode geometry and kinematics in unified neural fields\cite{noguchi2021neural}. Although many previous works focused on animals and human bodies to achieve natural motion deformation \cite{zhangLearningImplicitRepresentation2024a, jung2023deformable, lei2024gart}, there are still explorations on articulated objects.

\subsection{Articulated Object Manipulation}
The manipulation of articulated objects plays a fundamental role in a wide range of real-world scenarios and constitutes a critical task in the field of embodied intelligence. However, it remains highly challenging due to the complex physical properties inherent to such objects.
In the realm of articulated object manipulation, early research endeavors primarily centered around imitation learning~\cite{gu2023maniskill2, guhur2023instruction}. This methodology harnesses expert demonstrations to formulate manipulation policies. However, imitation learning is beset with distribution shift problems. Furthermore, the collection of diverse and comprehensive demonstrations poses a formidable challenge, being labor intensive in terms of both time consumption and substantial costs. Lately, a series of visual perception approaches~\cite{jain2021screwnet, wang2024articulated, geng2023gapartnet}, have been designed to estimate instance-level or category-level articulation parameters which in turn enables the generation of manipulation trajectories. VAT-Mart~\cite{wu2021vat}, Where2Act~\cite{mo2021where2act}, UMPNet~\cite{xu2022universal} are capable of predicting open-loop task-specific motion trajectories, facilitating the execution of object manipulation operations. However, open-loop operations have drawbacks, such as lack of real-time feedback and sensitivity to initial parameters and models, restricting their use in complex scenarios.
By contrast, optimal control methods~\cite{mittal2022articulated, yu2023gamma} directly integrate articulation models as constraints. They allow for the optimization of manipulation trajectories, leveraging the recognized joint types and parameters. Nevertheless, the presence of inaccuracies in parameter estimation can result in failures, undermining the reliability and effectiveness of the manipulation policy.

\subsection{Gaussian Splatting in Robotic Manipulation}

With the rapid development in 3D reconstruction techniques, 3DGS\cite{kerbl20233d} leverages explicit Gaussian points to represent the scene, enabling highly detailed and differential rendering. Gaussian Splatting achieves higher effectiveness and efficiency compared with implicit representations such as Neural Radiance Fields (NeRF)\cite{mildenhall2021nerf} with fast inference, high fidelity, and strong editability for novel view synthesis.  Recently, many studies have focused on leveraging 3DGS to bridge the gap between embodied simulators and the real world in the context of robotic manipulation\cite{qureshi2024splatsim, lou2024robo, lu2024manigaussian, zheng2024gaussiangrasper}. For example, Differentiable Robot Rendering\cite{liu2024differentiable} reconstructs robotic arms in GS and applies them to downstream tasks (e.g., robot pose estimation, text-to-robot hand gestures, etc.). However, its reliance on implicit linear blend skinning and appearance deformation often struggles with out-of-distribution scenarios. Additionally, the approach optimizes all arm links as a single entity, resulting in rendered images that lack finer details. For instance, the gripper is often rendered with high-frequency artifacts and dilation, posing a critical limitation for manipulation tasks. Similarly, RoboGSim\cite{li2024robogsim} reconstructs both the robotic arms and objects in the scene, creating digital twin assets of the objects and training policies for Sim2Real tasks within a 3DGS-based simulator. However, it offers little insight into articulated objects, which significantly limits its practical applicability.

\section{Preliminary} \label{Sec: prelim}

We model an articulated object $M$, which consists of $K$ movable parts, as $M=\{m_i\}_{i=1}^K$, along with their corresponding joints parameters $J=\{\psi_i=(\mathbf{u}_i, \mathbf{q}_i, c_i)\}_{i=1}^K$, where $\mathbf{u}_i\in\mathbb{R}^3$ is a unit vector representing the joint axis, $\mathbf{q}_i\in\mathbb{R}^3$ denotes its origin and $c_i$ is the joint type.
The object $M$ is observed through multi-view RGB-D images $\{I_i\in\mathbb{R}^{4\times H\times W}\}_{i=1}^N$. 
3DGS takes these multi-view images as input and outputs a scene $\mathcal{G}=\{g_i\}_{i=1}^n$ represented by $n$ Gaussian spheres, each Gaussian sphere has a tuple of attributes $(\mu_i, r_i, s_i, o_i, c_i, W_i)$. 
Here, $\mu\in\mathbb{R}^3$, $r\in SO(3)$, $s\in\mathbb{R}^{3+}$, $o\in\mathbb{R}$ and $c\in SH(3)$ denote the mean, rotation factor, scaling factor, opacity and color defined by spherical harmonic coefficients. 
The learnable skinning weights $W\in\mathbb{R}^{K+1}$ quantify the association between each Gaussian and its corresponding articulation part, and are subsequently employed in dynamic articulation modeling and optimization during manipulation.
\section{Method} \label{Sec: method}

\begin{figure*}[]
  \begin{center}
   \includegraphics[width=0.98\linewidth]{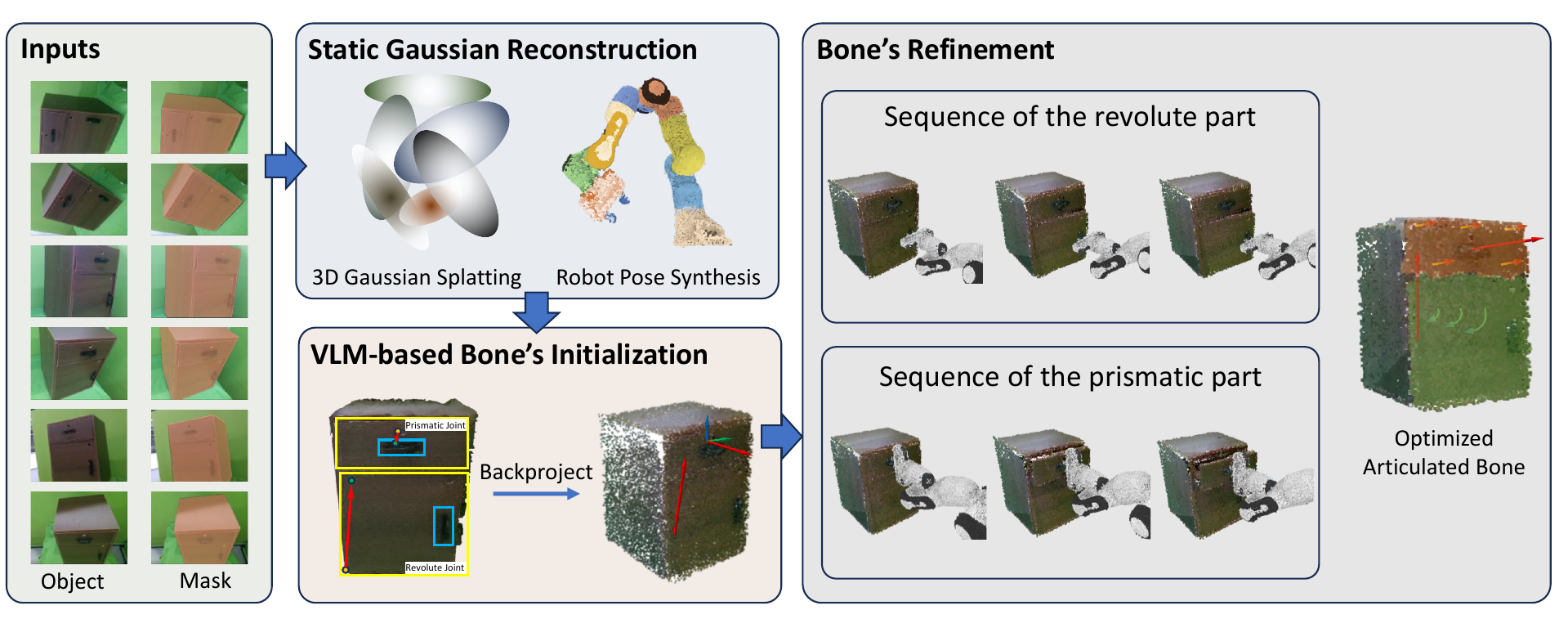}
  \end{center} 
  \caption{\textbf{Pipeline of ArtGS.} Starting from multi-view RGB-D inputs and object masks, ArtGS performs static 3DGS reconstruction and synthesizes robot poses. It then uses a VLM-based bone initialization module to infer the articulated bone parameters through visual-language reasoning. Finally, the Bone Refinement module dynamically optimizes the parameters of revolute and prismatic parts, producing a precise kinematic model of the articulated objects.}
  \label{Fig: pipeline}
\end{figure*}

As shown in Fig.~\ref{Fig: pipeline}, ArtGS comprises three key components: Static Gaussian Reconstruction for articulated objects and robot poses, VLM-based Articulated Bone Initialization for joint parameter estimation, and Dynamic 3D Gaussian Articulation Modeling for optimizing the parameters of the Articulated Bone.

\subsection{Static 3D Gaussian Reconstruction} \label{Sec: method1}

We employ the 3D Gaussian Splatting method to reconstruct both the articulated objects in their static state and the robotic arm in a specific pose. The static scene is represented as a collection of 3D Gaussian spheres, each of which can be expressed as a probability density function:
\begin{equation}
f_g = \mathcal{N}(\mu, \mathbf{R}\mathbf{S}\mathbf{S}^\mathsf{T}\mathbf{R}^\mathsf{T}),
\end{equation}
where $\mathbf{R}\mathbf{S}\mathbf{S}^\mathsf{T}\mathbf{R}^\mathsf{T}$ computes the full covariance matrix, $\mathbf{R}$ denotes the rotation matrix, and $\mathbf{S}$ denotes the scaling matrix.
When rendering under a given camera's extrinsic matrix $C_{ext}$ and intrinsic matrix $C_{int}$, the perspective projection of a 3D Gaussian onto the camera's image plane can be approximated as a 2D Gaussian:
\begin{equation}\label{gs_eq2}
\mu_{2D} = \pi(\mu; C_{ext}, C_{int}), \quad \Sigma_{2D} = P_{J}C_{ext} \Sigma C_{ext}^\mathsf{T}P_{J}^\mathsf{T},
\end{equation}
where $P_J$ is the Jacobian of the perspective projection. 
For each pixel, its color is determined by the ordered set of Gaussians  $\mathcal{G}_c\subset\mathcal{G}$ that overlap at the pixel:
\begin{equation}
C=\sum_{g_i\in\mathcal{G}_c}c_i\alpha_i \prod_{j=1}^{i-1}(1-\alpha_i).
\end{equation}
Here, $c_i$ and $\alpha_i$ represent the color and density contributions of Gaussian $g_i$ at this pixel, computed using learnable per-point opacity and SH color coefficients.

For the reconstruction of a \textbf{robotic arm}, since we have access to the URDF file, each joint and link can be treated as a static scene, similar to the reconstruction approach discussed above. By rendering and optimizing each link individually, this method effectively preserves fine details, addressing the issue of detail loss in distant regions that often arises in holistic optimization approaches. Given a pose $\mathbf{p}$, the transformation matrices for each joint can be easily constructed using the Modified Denavit-Hartenberg (MDH) convention\cite{corke2007simple}, a parameterized model widely used to describe the kinematic chain of robotic manipulators.
Using MDH parameters $\Theta_i=(\beta_i,a,d_i,\theta_i)$, the twist angle, link length, link offset and the joint angle, the coordinate frame of joint $i$ with respect to joint $i-1$, $A_i$ can be written as:
\begin{equation}
\begin{bmatrix}
\cos\theta_i & -\sin\theta_i\cos\beta_i & \sin\theta_i\sin\beta_i & a_i\cos\theta_i\\
\sin\theta_i & \cos\theta_i\cos\beta_i & -\cos\theta_i\sin\beta_i & a_i\sin\theta_i\\
0 & \sin\beta_i & \cos\beta_i & d_i\\
0 & 0 & 0 & 1
\end{bmatrix}.
\end{equation}
By sequentially multiplying these matrices, we can obtain the transformation matrix $T_j(\mathbf{p})$ of joint $j$ with respect to the base joint $0$ is produced as below:
\begin{equation}
    T_j(\mathbf{p}) = \prod_{i=0}^{j-1} A_i.
\end{equation}

We further apply the Gaussians of each link with their corresponding joint transformation matrices, achieving forward-kinematic-driven control of the robot's Gaussian points. 


\subsection{VLM-based Bones Reasoning} \label{Sec:method2}

In this module, we leverage a fine-tuned Visual-Language Model (VLM), based on InternVL-2.5-4B\cite{chen2024internvl}, to initialize the joint parameters estimation process. By utilizing the novel view synthesis capability of 3DGS, we can generate a frontal view of the object using the constructed Gaussian Points. This allows us to observe all movable parts, thereby enhancing the stability of the Visual Question Answering (VQA) results.

For any articulated object $M$, we observe it through an RGB image $I^{rgb}$ and a depth map $I^{d}$. Since an articulated object may contain multiple manipulable parts, we first use the fine-tuned VLM to segment all movable components and infer their joint parameters. We formulate the inference results as a list of pairs $(\mathcal{B}_i, c_i)$, where $\mathcal{B}_i\in \mathbb{R}^{2\times 2}$ represents the bounding box (BBOX) of the part in the image, and $c_i$ indicates the joint type.

For each part, we use the predicted bounding box as a visual prompt, enabling the model to focus specifically on this part. The queried joint parameters are defined by two vertices $(x_1, y_1, x_2, y_2)$ in pixel coordinates. Using the provided depth map, these two points can be mapped to the world coordinate as $(\mathbf{p}_1, \mathbf{p}_2)$. 

If the predicted joint type is \textbf{revolute}, we sample points along the line connecting $\mathbf{p}_1$ and $\mathbf{p}_2$. We then perform Principal Component Analysis (PCA) on these sampled points, discarding eigenvalues that are either excessively large or disproportionately small. This process allows us to focus on the most significant eigenvalue, from which we can derive the principal direction of the joint axis. Additionally, we take the mean position of the sampled points as the base point of the axis.
If the predicted joint type is \textbf{prismatic}, the two predicted vertices may be too close to each other, leading to potential numerical computation issues. Since we are only interested in the direction of the translational axis, we sample one direction along the horizontal and vertical lines of the predicted bounding box, respectively. The cross product of these two directions is then used as the final result.

The parameter estimation for all joints, formulated as $J_{init}=\{\psi_{i}\}_{i=1}^K$, guides the subsequence impedance control to manipulate each movable part of the articulated object. 
To facilitate manipulation, we also introduce an auxiliary task of identifying operable handles, which also enhances the spatial reasoning capabilities of our fine-tuned VLM.
More details about the design of Visual Question Answering (VQA) prompts are given in Table \ref{tab:VQA_Task}. 

\begin{table*}[!htbp]
\caption{Overview of the Articulated Bone tasks.}

\centering
\renewcommand\arraystretch{1.5}
\resizebox{1\textwidth}{!}{
\begin{tabular}{c|c|l}
\hline

\textbf{Capabilities} &
\textbf{Tasks} &
\multicolumn{1}{c}{\textbf{Examples of Task Templates}} \\ \hline 

\multicolumn{1}{l|}{Object Classification.} &
\multicolumn{1}{l|}{Classification} &
\begin{tabular}{@{}l@{}}
    \textcolor{blue}{\texttt{User}}: What is the object, and what are its movable components? \\ 
    \textcolor{customgreen}{\texttt{VLM}}: ~ This is a \textbf{object category} with \textbf{number of movable parts}.
\end{tabular}  \\ \hline 

\multicolumn{1}{l|}{Articulated Object Understanding.} &
\multicolumn{1}{l|}{Bones Reasoning} &
\begin{tabular}{@{}l@{}}
    \textcolor{blue}{\texttt{User}}: What are the joint types, joint parameters, and specific bounding box coordinates of these movable parts?\\ 
    \textcolor{customgreen}{\texttt{VLM}}: ~ Part 1 is a \textbf{door}, and its \textbf{joint type} is revolute. The \textbf{joint parameter} is [[...]]. Its \textbf{bounding box} is [[...]].
\end{tabular}  \\ \hline

\multicolumn{1}{l|}{Interactive Part Grounding.} &
\multicolumn{1}{l|}{Part Detection} &
\begin{tabular}{@{}l@{}}
    \textcolor{blue}{\texttt{User}}: What are the bounding box coordinates of the \textbf{handle} for the movable part within <box>[[...]]</box>?\\ 
    \textcolor{customgreen}{\texttt{VLM}}: ~ Its handle is [[...]].
\end{tabular}  \\ \hline

\end{tabular}%
}
\label{tab:VQA_Task}
\vspace{-15pt}
\end{table*}


\subsection{Dynamic 3D Gaussian Articulation Modeling} \label{Sec:method3}

We employ impedance control to enable the manipulator to follow a desired trajectory $x_d$ while accounting for external force $F_{ext}$ resulting from the interaction between the robot and the environment. The desired trajectory $x_d$ is computed based on the joint configurations $J$ of the articulated object. The dynamic model of impedance control is:
\begin{equation}
M(\ddot{x}_c-\ddot{x}_d)+D(\dot{x}_c-\dot{x}_d)+K(x_c-x_d)=F_{ext}
\end{equation}
where $M$ is the inertia matrix, $D$ is the damping matrix, $K$ is the stiffness matrix, and $[\ddot{x}_c, \dot{x}_c, x_c]$ is the outputs of end-effector’s trajectory.

After manipulation, we leverage differential rendering to integrate the 3D Gaussian representation $\mathcal{G}$ with a rigid skeleton $\mathcal{B}$ for dynamic reconstruction and refinement phase, enabling precise articulation modeling.

The rigid skeleton $\mathcal{B}$ of an articulated object with $K$ joints generates a list of special Euclidean transformations based on the pose $\theta$:
\begin{equation}
\begin{bmatrix}
    B_0, B_1, B_2, \dots, B_K
\end{bmatrix} = \mathcal{B}(\theta),
\end{equation}
where $B_i \in SE(3)$ represents the rigid transformation that maps the canonical joint coordinate frame to the articulated one, depending on the corresponding joint type $c_i$. If the joint type is revolute, $B_i$ is derived from the Rodrigues formula, which calculates a rotation matrix from the axis-angle representation. In contrast, if the joint type is prismatic, $B_i$ is obtained by the translation along the joint axis. $B_0$ represents the identity transformation in SE(3), which corresponds to a stationary state and is used to distinguish between movable parts and the base. We initialize the rigid skeleton using priors of joint type and parameters estimation obtained from our fine-tuned VLM. 

The mean $\mu$ and rotation factor $r$ of each Gaussian point in the canonical space can be deformed to the articulated space via the linear blend skinning (LBS) using point-wise learnable skinning weight. More specifically, the transformation result of a Gaussian point $g_i$ is computed as:
\begin{equation}
    \mu_i' = \left(\sum_{j=0}^K W_{ij}B_j\right)\mu_i,
    \quad
    r_i' = \left(\sum_{j=0}^K W_{ij}B_j\right)r_i.
\end{equation}

In the optimization phase, using the multi-view RGB-D images $\{I_t\in \mathbb{R}^{4\times H\times W\times N}\}_{t=1}^T$ collected during manipulation and the rendered images and depth maps, $\{I^{rgb}_t(\mathcal{G}', \mathcal{G}_{robot})\}$ and $\{I^{d}_t(\mathcal{G}', \mathcal{G}_{robot})\}$, we minimize the loss function to learn the joint parameters $J$ and object's state $\{\theta_t\}_{t=1}^T$:
\begin{equation}
\min_{J=\{\psi_i\}_{i=1}^K, \{\theta_t\}_{t=1}^T} \mathcal{L}=\lambda_{L1} L_1+\lambda_{SSIM}L_{SSIM}+L_{reg}.
\end{equation}
 
By iteratively planning trajectories and optimizing articulation parameters based on the actual manipulation trajectories, we can continuously improve the modeling accuracy of the articulated objects.
\section{Experiment} \label{Sec: experiment}

In this section, we conduct comprehensive experiments on modeling and manipulation in both simulated and real-world environments. We first perform quantitative evaluations on joint parameters estimation across various interactable categories, showing that our method can robustly and accurately model the articulations. We then evaluate the manipulation performance to show that the estimated joint parameters can effectively guide the manipulator in executing tasks.

\begin{figure}[t!]
  \begin{center}
   \includegraphics[width=1\linewidth]{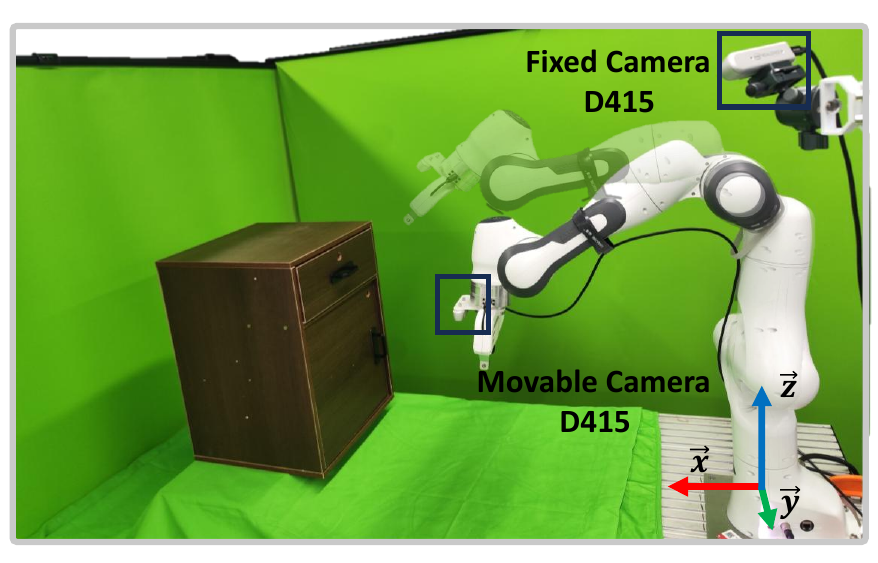}
  \end{center}
  \vspace{-15pt}
      \caption{\textbf{Settings of Real-world Environment}. Two cameras are used: an eye-in-hand camera for capturing multi-view images in static 3D GS reconstruction and a fixed third-person camera for single-view images to update the dynamic 3D GS. }
\label{Fig: setting}
\vspace{-15pt}
\end{figure}

{\bf Environments.}We carry out simulation experiments within the SAPIEN simulator~\cite{xiang2020sapien}, which is specifically designed to offer high-fidelity physical simulations for robots, rigid bodies, and articulated objects. It also boasts remarkable photo-realistic rendering capabilities, which play a crucial role in bridging the gap between simulation and the real world, thereby promoting effective sim-to-real generalization.

Furthermore, we conduct real-world experiments utilizing a 7-Dof Franka robot, as shown in Fig.~\ref{Fig: setting}. A RealSense D415 RGB-D camera is mounted on the wrist of the Franka robot to precisely capture the point clouds of all the articulated objects involved in the experiment for reconstruction. To track the manipulation process, an additional D415 camera is installed on the left side of the robot's operation platform, which has been meticulously calibrated with respect to the robot's base coordinate system, ensuring that the data acquisition and subsequent processing are both accurate and highly reliable.

Then we implement \textbf{ArtGS} to perform manipulation tasks on five distinct real-world articulated objects including storage (prismatic joints), drawer (revolute joints), cabinet (revolute and prismatic joints), and microwave (revolute joints).

{\bf Datasets.}
Based on the GAPartNet dataset \cite{geng2023gapartnet}, we have organized and selected 7 categories of 100 articulated objects from it. These categories are: storage (revolute joints), drawer (prismatic joints), cabinet (revolute and prismatic joints), microwave (revolute joints), oven (revolute joints), refrigerator (revolute joints), and dishwasher (revolute joints).
We select these articulated objects because they have pre-defined handles that facilitate the manipulation tasks through interaction. Furthermore, the data for each category is partitioned into "train" and "evaluation" subsets, enabling training and assessment of the model's generalization capability. We fine-tuned the base model, InternVL-2.5-4B, using a dataset of 14k text-image pairs.

\subsection{Articulated Object Modeling} \label{Sec: experiment1}

\textbf{Baselines.}
We compare our proposed method with three different baselines under the identical setting.
\begin{itemize}
\item \textbf{ANCSH}\cite{li2020category} represents a cutting-edge approach for articulated object modeling. Initially, it segments the object into articulated parts using single-view point clouds. Subsequently, it transforms the points into a normalized coordinate space to estimate the joint parameters.

\item \textbf{GAMMA}\cite{yu2023gamma} first introduces the concept of cross-category articulated objects. It directly conducts joint parameter estimation within the original point cloud space. By learning the articulation modeling from a wide variety of articulated objects belonging to different categories, GAMMA enables more generalized joint parameter estimation.

\item \textbf{Ditto}\cite{jiang2022ditto} is a method designed to handle category-agnostic 3D object manipulation. Ditto learns from diverse object categories to predict category-specific kinematic properties, including joint types and their parameters. We conducted tests on Ditto to model the articulated object, using a 30° and 30 cm configuration.
\end{itemize}

\textbf{Results.}
As shown in Table~\ref{Tab: joint exp}, ANCSH~\cite{li2020category}, which relies on category-level priors, performs the worst overall, yielding the highest joint axis (AE) and origin errors (OE) across most object categories. GAMMA exhibits moderate performance, with AE ranging from 9.23° to 12.67° and OE between 8.96 cm and 11.94 cm. Compared to single-frame point clouds, utilizing two consecutive frames enables more effective capture of motion information, thereby significantly enhancing the joint parameter estimation accuracy achieved by Ditto. In contrast, \textbf{ArtGS} consistently outperforms all other methods across various objects, achieving the lowest AE and OE values, which underscores its superior robustness and accuracy in joint parameter estimation. In contrast, \textbf{ArtGS} consistently outperforms all other methods across various objects, achieving the lowest AE and OE values. It combines accurate joint parameter estimation with interactive optimization, enabling robust and adaptive manipulation across diverse articulated objects.

\begin{table*}[htbp]
\caption{Results of articulated object joint parameter estimation (Joint estimation error  $\downarrow$)}
\centering
\renewcommand\arraystretch{1.2}
\begin{threeparttable}
\resizebox{1\textwidth}{!}{
\begin{tabular}{c|ccccccccccccc}
\toprule 
\multicolumn{1}{c|}{\multirow{2}{*}{Method}} & \multicolumn{2}{c}{Dishwasher} & \multicolumn{2}{c}{Refrigerator} & \multicolumn{2}{c}{Oven} & \multicolumn{2}{c}{Microwave}& \multicolumn{2}{c}{Storage}& Drawer & \multicolumn{2}{c}{Cabinet} \\
\multicolumn{1}{c|}{} 
& AE & OE 
& AE & OE 
& AE & OE 
& AE & OE 
& AE & OE 
& AE 
& AE & OE \\
\hline

 ANCSH~\cite{li2020category} & 15.32 & 9.26  & 16.34 & 8.76 & 14.45 & 10.30 & 15.36 & 8.41 & 14.43 & 9.70 &13.62 & 13.17 & 10.53 

\\

 GAMMA~\cite{yu2023gamma} & 12.67 & 9.60 & 10.42 & 9.97 & 11.13 & 8.96 &11.96 & 9.67 &9.23 & 10.02 & 9.65 & 10.24 & 11.94

\\
 Ditto~\cite{jiang2022ditto}& 
3.63 & 4.70 & \textbf{0.46} & 9.53 & 5.05 & 7.43 &
7.94 & 6.58 & \textbf{0.31} & 4.69 & \textbf{5.06} &  \textbf{5.88} & 9.67
\\

ArtGS (w/o opt.) &
20.90 & 5.91 & 25.01 & 4.81 & 15.62 & 4.23 &
12.05 & 14.68 & 15.92 & 5.32 & 20.98 & 19.00 & 15.63
\\

 ArtGS &
\textbf{3.01} & \textbf{2.17} & 1.10 & \textbf{2.86} & \textbf{1.53} & \textbf{3.91} &
\textbf{3.17} & \textbf{2.68} & 1.52 & \textbf{2.23} & 8.32 & 6.50 & \textbf{5.33}

\\

\toprule 
\end{tabular}
}
\begin{tablenotes}
\scriptsize
\item In the evaluation of cabinet-like objects, Ditto exhibits a failure rate of 3 out of 109 (marked with *). 
AE= joint axis error  $/ ^{\circ}$; OE = joint origin error /cm.
\end{tablenotes}
\end{threeparttable}
 \label{Tab: joint exp}
 \vspace{-20pt}
\end{table*}

\subsection{Articulation Manipulation} \label{Sec: experiment2}

\textbf{Baselines.}
We compare our proposed method with five different baselines under the identical setting.
\begin{itemize}
\item \textbf{TD3}~\cite{dankwa2019twin} serves as a widely adopted baseline for robot manipulation tasks. In this approach, the observations encompass point clouds and the states of the end-effector. The defined action corresponds to the incremental alterations in the state of the end-effector.

\item \textbf{Where2Act}~\cite{mo2021where2act} identifies grasping points with higher actionability for manipulation tasks and estimates a short-term manipulation action for each of these points.

\item\textbf{VAT-Mart}~\cite{wu2021vat} leverages 3D object-centric actionable visual priors to predict interaction-aware and task-aware visual action affordances, along with trajectory proposals, for manipulation tasks.

\item\textbf{UMPNet}~\cite{xu2022universal} employs a policy network to infer closed-loop action sequences from a single image, enabling 6-DOF actions with flexible trajectory lengths to manipulate objects with various articulation structures. While the original method utilizes suction, we employ the flyhand gripper in our approach.

\end{itemize}

\textbf{Results.}
The table presents the success rates of various methods in simulating articulated object manipulation. Performance varies significantly across different objects, such as dishwashers, refrigerators, and cabinets.
TD3 demonstrates low success rates, ranging from 3.1\% to 6.6\%. Where2Act shows a slight improvement, with success rates between 8.4\% and 11.2\%. VAT-Mart achieves further enhancements, with rates ranging from 18.6\% to 40.9\%. UMPNet exhibits better performance, with success rates from 30.2\% to 39.2\%, while GAMMA shows a more substantial improvement, ranging from 42.1\% to 56.8\%.
Our method, \textbf{ArtGS}, which achieves the highest success rate in each category, with success rates from 62.4\% to 90.3\%, outperforms all other approaches. This highlights the superior effectiveness of our method in articulated object modeling and manipulation through visual-physical modeling.
\begin{table*}[htbp]
\caption{Simulation articulated object manipulation results (Success rate \%   $\uparrow$)}
\centering
\renewcommand\arraystretch{1.2}
\begin{threeparttable}
\resizebox{0.9\textwidth}{!}{
\begin{tabular}{c|ccccccc}
\toprule 
\multicolumn{1}{c|}{\multirow{2}{*}{Method}} & \multicolumn{1}{c}{Dishwasher} & \multicolumn{1}{c}{Refrigerator} & \multicolumn{1}{c}{Oven} & \multicolumn{1}{c}{Microwave}& \multicolumn{1}{c}{Storage}& Drawer & \multicolumn{1}{c}{Cabinet} \\
\multicolumn{1}{c|}{\multirow{2}{*}{}} & \multicolumn{1}{c}{revolute} & \multicolumn{1}{c}{revolute} & \multicolumn{1}{c}{revolute} & \multicolumn{1}{c}{revolute}& \multicolumn{1}{c}{revolute}& prismatic & \multicolumn{1}{c}{revolute/prismatic} \\
\hline
      
TD3~\cite{dankwa2019twin}   & 5.6 & 4.6 & 4.3 & 3.1 & 4.9 & 5.8 & 6.6 \\
Where2Act~\cite{mo2021where2act}  &9.2 & 10.1 & 8.4 & 8.7 & 9.8& 11.2 & 11.0 \\
VAT-Mart~\cite{wu2021vat}   &24.3 & 19.9 & 21.3 & 20.4 & 18.6 & 40.9 & 34.5 \\
UMPNet~\cite{xu2022universal} &30.9 &35.7 & 33.6 & 37.8 & 30.2 & 39.2 & 36.4\\
GAMMA~\cite{yu2023gamma}  & 49.2 & 42.1 & 50.8 & 47.5 & 50.2 & 56.8 & 55.3 \\
ArtGS (w/o opt.) & 45.0 & 40.6 & 39.5 & 37.3 & 38.8 & 43.7 & 46.1 \\
ArtGS & \textbf{62.4} & \textbf{75.2} & \textbf{67.9} & \textbf{79.1} & \textbf{78.4} & \textbf{90.3} & \textbf{76.8} \\
\toprule 
\end{tabular}
}
\begin{tablenotes}
\item 
\end{tablenotes}
\end{threeparttable}
 \label{Tab: manip}
 \vspace{-25pt}
\end{table*}

\subsection{Ablation Experiment} \label{Sec: ablation}
Although it is necessary to use fine-tuned VLM to provide an initial set of joint parameters for our method, the optimization capability of our model is not constrained by the inherent limitations of the VLM itself. As demonstrated in the last two rows of Table~\ref{Tab: joint exp} and Table~\ref{Tab: manip}, when the VLM accurately predicts the joint type and axis, our method achieves superior optimization results in the majority of cases. Conversely, even in scenarios where the VLM produces significant errors in axis estimation (e.g., the Joint Axis Error exceeds 20° for the dishwasher, refrigerator, and drawer), \textbf{ArtGS} optimizes the joint parameters by leveraging the differentiable rendering of 3DGS to ensure spatial and temporal consistency, enabling efficient interactive optimization, which significantly improves the success rate of manipulation tasks for articulated objects.

\subsection{Cross-Embodiment Experiment} \label{Sec: experiment3}

\begin{figure}[t]
  \begin{center}
   \includegraphics[width=1\linewidth]{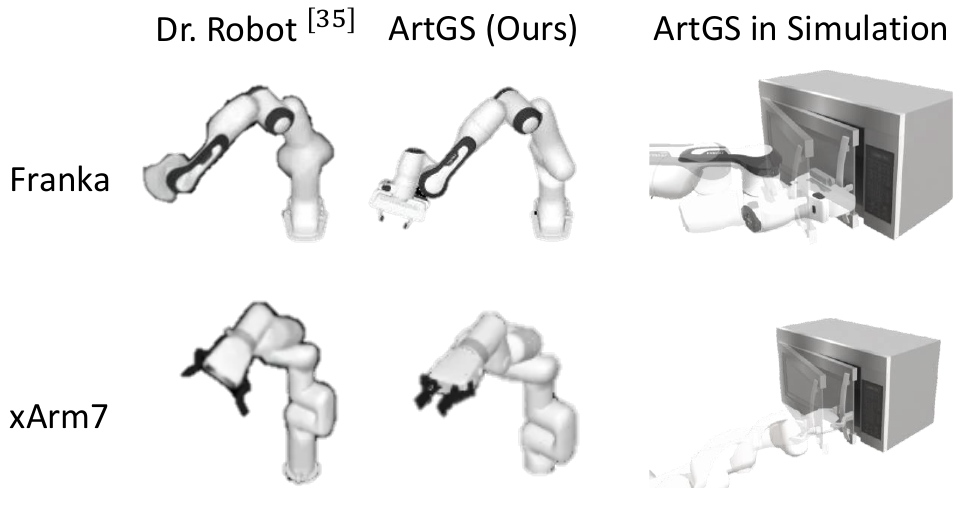}
  \end{center}
  \vspace{-20pt}
      \caption{\textbf{Cross-Embodiment Experiment.} This figure demonstrates the cross-embodiment capability of ArtGS. The first and second rows show qualitative results for the Franka and xArm7 robotic arms, respectively. The first column displays the robotic arm reconstruction results from Dr. Robot\cite{liu2024differentiable}, the second column presents our higher-quality digital assets, and the third column illustrates the manipulation results of ArtGS in a simulated environment across different robotic arms.}
\label{Fig: cross}
\end{figure}
Building upon our outstanding proficiency in 3DGS modeling for articulated objects, we can execute highly accurate reconstructions of diverse robotic arms, endowing these robotic arms with remarkable Cross-Embodiment capabilities.

As illustrated in Fig.~\ref{Fig: cross}, we have presented the reconstruction outcomes and the visual effects within the simulation environment for both the Franka and xArm7 robotic arms. Additionally, we have juxtaposed these results with those of the robotic arm reconstruction by Dr. Robot~\cite{liu2024differentiable}.
Upon careful observation, it becomes evident that our reconstruction surpasses the alternatives significantly. In particular, the rendering of the crucial gripper component is of superior quality. This meticulous attention to detail in the gripper's rendering is of paramount importance for the generation of high-fidelity simulation data about grasp, ensuring that the simulated scenarios closely mirror real-world conditions.
\subsection{Real World Experiment} \label{Sec: experiment4}

\begin{table}[t]
\caption{Real-world Manipulation Results (Success rate \% $\uparrow$)}
\vspace{-20pt}
\begin{center}
\renewcommand\arraystretch{1.2}
\begin{threeparttable}
\resizebox{0.5\textwidth}{!}{
\begin{tabular}{c|ccccc}
\toprule
    \multirow{1}{*}{~}
    & \multicolumn{1}{c}{cabinet(drawer)} & \multicolumn{1}{c}{cabinet(door)} & \multicolumn{1}{c}{drawer} & \multicolumn{1}{c}{storage} & \multicolumn{1}{c}{microwave} \\
    \hline 
ArtGS (w/o opt.) &6/10 & 4/10 & \multicolumn{1}{c}{7/10} & 5/10 &  4/10 \\

ArtGS & \textbf{9/10}  & \textbf{7/10} & \multicolumn{1}{c}{\textbf{10/10}} & \textbf{{9/10}} & \textbf{8/10} \\

\bottomrule
\end{tabular}
}
\begin{tablenotes}
\item 
\begin{flushleft}
\small
For each object, we conducted 10 trials, randomly positioning \\ them within the robot's workspace, followed by the execution \\ of modeling and manipulation tasks.
\end{flushleft}

\end{tablenotes}
\end{threeparttable}
\end{center}
\label{tab:real_manipulation}
\vspace{-25pt}
\end{table}

\begin{figure}[t]
  \begin{center}
   \includegraphics[width=1\linewidth]{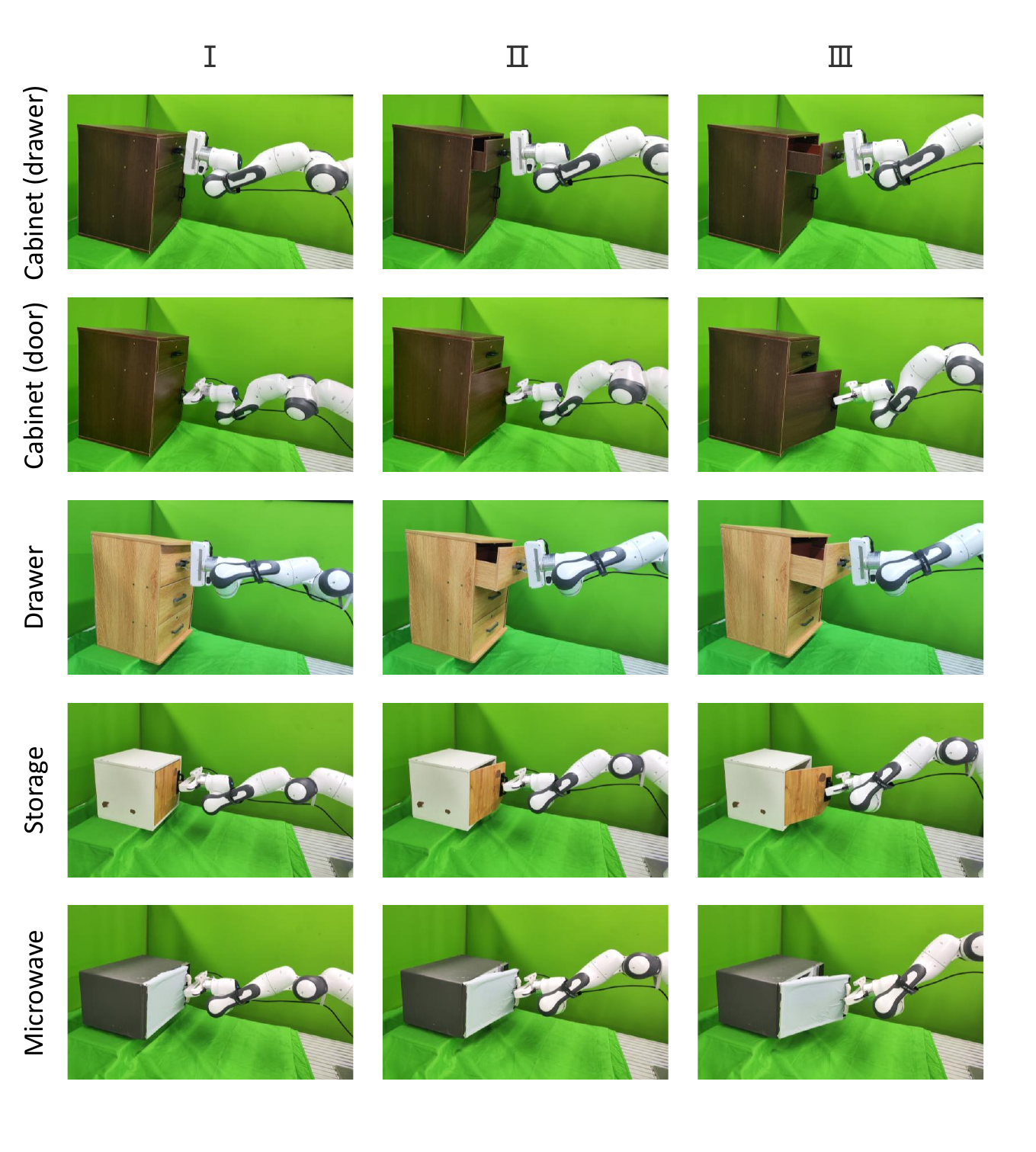}
  \end{center}
  \vspace{-20pt}
      \caption{\textbf{Real-world Experiment.} We implement ArtGS and only fine-tuned VLM in the real-world experiments. Manipulation tasks include opening the door of cabinet (2 parts), drawer, storage, and microwave}
\label{Fig: real}
\vspace{-20pt}
\end{figure}

We apply our method to real-world objects to evaluate its generalization capability. In this setup, the robot uses a 7-DOF Franka arm, and we calibrate both the in-hand camera and a fixed third-person-view camera to the base coordinate system of the robotic arm. During the initial reconstruction phase, we move the in-hand camera across 15 different viewpoints using the Franka arm to complete the static articulation reconstruction. As shown in Fig.~\ref{Fig: real} and Table~\ref{tab:real_manipulation}, through 3DGS reconstruction and articulated bone parameter optimization, we significantly enhance the \textbf{ArtGS}’s generalization ability, resulting in excellent performance in both object modeling and manipulation tasks. Due to space constraints, detailed
information is provided in the \href{https://sites.google.com/view/artgs/home}{website}.

\section{Conclusion} \label{Sec: conclusion}

This paper proposes \textbf{ArtGS}, a framework that transforms 3D Gaussian Splatting from a high-fidelity reconstruction tool into a visual-physical model for articulated objects. By integrating articulated bones with 3DGS representations, we enable both the modeling and interactive optimization of articulated objects, fully leveraging the spatiotemporal consistency of dynamic 3DGS and differentiable rendering to enhance optimization efficiency. A key feature of our approach is the dynamic 3D Gaussian Splatting, which guarantees differentiable manipulation trajectories while maintaining both temporal and spatial consistency. These properties not only improve optimization efficiency but also address occlusion challenges during manipulation. Furthermore, \textbf{ArtGS} demonstrates robust cross-embodiment adaptability, seamlessly supporting various robotic arms (e.g., Franka and xArm7), expanding its applicability across diverse robotic systems. Experimental results in both simulated and real-world settings show superior performance compared to existing methods. Looking ahead, we aim to extend this framework to address more complex scenes.

\bibliographystyle{IEEEtran}
\bibliography{references}

\begin{thebibliography}{10}
\providecommand{\url}[1]{#1}
\csname url@samestyle\endcsname
\providecommand{\newblock}{\relax}
\providecommand{\bibinfo}[2]{#2}
\providecommand{\BIBentrySTDinterwordspacing}{\spaceskip=0pt\relax}
\providecommand{\BIBentryALTinterwordstretchfactor}{4}
\providecommand{\BIBentryALTinterwordspacing}{\spaceskip=\fontdimen2\font plus
\BIBentryALTinterwordstretchfactor\fontdimen3\font minus \fontdimen4\font\relax}
\providecommand{\BIBforeignlanguage}[2]{{%
\expandafter\ifx\csname l@#1\endcsname\relax
\typeout{** WARNING: IEEEtran.bst: No hyphenation pattern has been}%
\typeout{** loaded for the language `#1'. Using the pattern for}%
\typeout{** the default language instead.}%
\else
\language=\csname l@#1\endcsname
\fi
#2}}
\providecommand{\BIBdecl}{\relax}
\BIBdecl

\bibitem{wu2021vat}
R.~Wu, Y.~Zhao, K.~Mo, Z.~Guo, Y.~Wang, T.~Wu, Q.~Fan, X.~Chen, L.~Guibas, and H.~Dong, ``Vat-mart: Learning visual action trajectory proposals for manipulating 3d articulated objects,'' \emph{arXiv preprint arXiv:2106.14440}, 2021.

\bibitem{mu2025robotwin}
Y.~Mu, T.~Chen, Z.~Chen, S.~Peng, Z.~Lan, Z.~Gao, Z.~Liang, Q.~Yu, Y.~Zou, M.~Xu \emph{et~al.}, ``Robotwin: Dual-arm robot benchmark with generative digital twins,'' in \emph{Proceedings of the Computer Vision and Pattern Recognition Conference}, 2025, pp. 27\,649--27\,660.

\bibitem{yu2025forcevla}
J.~Yu, H.~Liu, Q.~Yu, J.~Ren, C.~Hao, H.~Ding, G.~Huang, G.~Huang, Y.~Song, P.~Cai \emph{et~al.}, ``Forcevla: Enhancing vla models with a force-aware moe for contact-rich manipulation,'' \emph{arXiv preprint arXiv:2505.22159}, 2025.

\bibitem{yu2023gamma}
Q.~Yu, J.~Wang, W.~Liu, C.~Hao, L.~Liu, L.~Shao, W.~Wang, and C.~Lu, ``Gamma: Generalizable articulation modeling and manipulation for articulated objects,'' in \emph{2024 IEEE International Conference on Robotics and Automation (ICRA)}.\hskip 1em plus 0.5em minus 0.4em\relax IEEE, 2024, pp. 5419--5426.

\bibitem{wang2024rpmart}
J.~Wang, W.~Liu, Q.~Yu, Y.~You, L.~Liu, W.~Wang, and C.~Lu, ``Rpmart: Towards robust perception and manipulation for articulated objects,'' in \emph{2024 IEEE/RSJ International Conference on Intelligent Robots and Systems (IROS)}.\hskip 1em plus 0.5em minus 0.4em\relax IEEE, 2024, pp. 7270--7277.

\bibitem{kerbl20233d}
B.~Kerbl, G.~Kopanas, T.~Leimk{\"u}hler, and G.~Drettakis, ``3d gaussian splatting for real-time radiance field rendering.'' \emph{ACM Trans. Graph.}, vol.~42, no.~4, pp. 139--1, 2023.

\bibitem{liCategoryLevelArticulatedObject2020}
X.~Li, H.~Wang, L.~Yi, L.~J. Guibas, A.~L. Abbott, and S.~Song, ``Category-{{Level Articulated Object Pose Estimation}},'' in \emph{2020 {{IEEE}}/{{CVF Conference}} on {{Computer Vision}} and {{Pattern Recognition}} ({{CVPR}})}.\hskip 1em plus 0.5em minus 0.4em\relax IEEE, pp. 3703--3712.

\bibitem{bozicNeuralDeformationGraphs2021}
A.~Bozic, P.~Palafox, M.~Zollhofer, J.~Thies, A.~Dai, and M.~Niesner, ``Neural {{Deformation Graphs}} for {{Globally-consistent Non-rigid Reconstruction}},'' in \emph{2021 {{IEEE}}/{{CVF Conference}} on {{Computer Vision}} and {{Pattern Recognition}} ({{CVPR}})}.\hskip 1em plus 0.5em minus 0.4em\relax IEEE, pp. 1450--1459.

\bibitem{wu2025reartgs}
D.~Wu, L.~Liu, Z.~Linli, A.~Huang, L.~Song, Q.~Yu, Q.~Wu, and C.~Lu, ``Reartgs: Reconstructing and generating articulated objects via 3d gaussian splatting with geometric and motion constraints,'' \emph{arXiv preprint arXiv:2503.06677}, 2025.

\bibitem{zengMARSMultimodalActive2024}
H.~Zeng, P.~Zhang, C.~Wu, J.~Wang, T.~Ye, and F.~Li, ``{{MARS}}: {{Multimodal Active Robotic Sensing}} for {{Articulated Characterization}},'' in \emph{Proceedings of the {{Thirty-ThirdInternational Joint Conference}} on {{Artificial Intelligence}}}.\hskip 1em plus 0.5em minus 0.4em\relax International Joint Conferences on Artificial Intelligence Organization, pp. 1634--1642.

\bibitem{jiang2022ditto}
Z.~Jiang, C.-C. Hsu, and Y.~Zhu, ``Ditto: Building digital twins of articulated objects from interaction,'' in \emph{Proceedings of the IEEE/CVF Conference on Computer Vision and Pattern Recognition}, 2022, pp. 5616--5626.

\bibitem{Weng2024NeuralIR}
Y.~Weng, B.~Wen, J.~Tremblay, V.~Blukis, D.~Fox, L.~J. Guibas, and S.~T. Birchfield, ``Neural implicit representation for building digital twins of unknown articulated objects,'' \emph{2024 IEEE/CVF Conference on Computer Vision and Pattern Recognition (CVPR)}, pp. 3141--3150, 2024.

\bibitem{jain2021screwnet}
A.~Jain, R.~Lioutikov, C.~Chuck, and S.~Niekum, ``Screwnet: Category-independent articulation model estimation from depth images using screw theory,'' in \emph{2021 IEEE International Conference on Robotics and Automation (ICRA)}.\hskip 1em plus 0.5em minus 0.4em\relax IEEE, 2021, pp. 13\,670--13\,677.

\bibitem{zeng2021visual}
V.~Zeng, T.~E. Lee, J.~Liang, and O.~Kroemer, ``Visual identification of articulated object parts,'' in \emph{2021 IEEE/RSJ International Conference on Intelligent Robots and Systems (IROS)}.\hskip 1em plus 0.5em minus 0.4em\relax IEEE, 2021, pp. 2443--2450.

\bibitem{yu2024uniaff}
Q.~Yu, S.~Huang, X.~Yuan, Z.~Jiang, C.~Hao, X.~Li, H.~Chang, J.~Wang, L.~Liu, H.~Li \emph{et~al.}, ``Uniaff: A unified representation of affordances for tool usage and articulation with vision-language models,'' \emph{arXiv preprint arXiv:2409.20551}, 2024.

\bibitem{chuCommanddrivenArticulatedObject2023}
R.~Chu, Z.~Liu, X.~Ye, X.~Tan, X.~Qi, C.-W. Fu, and J.~Jia, ``Command-driven {{Articulated Object Understanding}} and {{Manipulation}},'' in \emph{2023 {{IEEE}}/{{CVF Conference}} on {{Computer Vision}} and {{Pattern Recognition}} ({{CVPR}})}.\hskip 1em plus 0.5em minus 0.4em\relax IEEE, pp. 8813--8823.

\bibitem{yu2025generalizable}
Q.~Yu, C.~Hao, X.~Yuan, L.~Zhang, L.~Liu, Y.~Huo, R.~Agarwal, and C.~Lu, ``Generalizable articulated object perception with superpoints,'' in \emph{ICASSP 2025-2025 IEEE International Conference on Acoustics, Speech and Signal Processing (ICASSP)}.\hskip 1em plus 0.5em minus 0.4em\relax IEEE, 2025, pp. 1--5.

\bibitem{geng2023gapartnet}
H.~Geng, H.~Xu, C.~Zhao, C.~Xu, L.~Yi, S.~Huang, and H.~Wang, ``Gapartnet: Cross-category domain-generalizable object perception and manipulation via generalizable and actionable parts,'' in \emph{Proceedings of the IEEE/CVF Conference on Computer Vision and Pattern Recognition}, 2023, pp. 7081--7091.

\bibitem{kawanaUnsupervisedPoseawarePart2022}
Y.~Kawana, Y.~Mukuta, and T.~Harada, ``Unsupervised {{Pose-aware Part Decomposition}} for {{Man-Made Articulated Objects}},'' in \emph{Computer {{Vision}} – {{ECCV}} 2022}, S.~Avidan, G.~Brostow, M.~Cissé, G.~M. Farinella, and T.~Hassner, Eds.\hskip 1em plus 0.5em minus 0.4em\relax Springer Nature Switzerland, vol. 13663, pp. 558--575.

\bibitem{noguchi2021neural}
A.~Noguchi, X.~Sun, S.~Lin, and T.~Harada, ``Neural articulated radiance field,'' in \emph{Proceedings of the IEEE/CVF International Conference on Computer Vision}, 2021, pp. 5762--5772.

\bibitem{zhangLearningImplicitRepresentation2024a}
H.~Zhang, F.~Li, S.~Rawlekar, and N.~Ahuja, ``Learning implicit representation for reconstructing articulated objects,'' \emph{arXiv preprint arXiv:2401.08809}, 2024.

\bibitem{jung2023deformable}
H.~Jung, N.~Brasch, J.~Song, E.~Perez-Pellitero, Y.~Zhou, Z.~Li, N.~Navab, and B.~Busam, ``Deformable 3d gaussian splatting for animatable human avatars,'' \emph{arXiv preprint arXiv:2312.15059}, 2023.

\bibitem{lei2024gart}
J.~Lei, Y.~Wang, G.~Pavlakos, L.~Liu, and K.~Daniilidis, ``Gart: Gaussian articulated template models,'' in \emph{Proceedings of the IEEE/CVF conference on computer vision and pattern recognition}, 2024, pp. 19\,876--19\,887.

\bibitem{gu2023maniskill2}
J.~Gu, F.~Xiang, X.~Li, Z.~Ling, X.~Liu, T.~Mu, Y.~Tang, S.~Tao, X.~Wei, Y.~Yao \emph{et~al.}, ``Maniskill2: A unified benchmark for generalizable manipulation skills,'' \emph{arXiv preprint arXiv:2302.04659}, 2023.

\bibitem{guhur2023instruction}
P.-L. Guhur, S.~Chen, R.~G. Pinel, M.~Tapaswi, I.~Laptev, and C.~Schmid, ``Instruction-driven history-aware policies for robotic manipulations,'' in \emph{Conference on Robot Learning}.\hskip 1em plus 0.5em minus 0.4em\relax PMLR, 2023, pp. 175--187.

\bibitem{wang2024articulated}
X.~Wang, T.~Chen, Q.~Yu, T.~Xu, Z.~Chen, Y.~Fu, Z.~He, C.~Lu, Y.~Mu, and P.~Luo, ``Articulated object manipulation using online axis estimation with sam2-based tracking,'' \emph{arXiv preprint arXiv:2409.16287}, 2024.

\bibitem{mo2021where2act}
K.~Mo, L.~J. Guibas, M.~Mukadam, A.~Gupta, and S.~Tulsiani, ``Where2act: From pixels to actions for articulated 3d objects,'' in \emph{Proceedings of the IEEE/CVF International Conference on Computer Vision}, 2021, pp. 6813--6823.

\bibitem{xu2022universal}
Z.~Xu, Z.~He, and S.~Song, ``Universal manipulation policy network for articulated objects,'' \emph{IEEE robotics and automation letters}, vol.~7, no.~2, pp. 2447--2454, 2022.

\bibitem{mittal2022articulated}
M.~Mittal, D.~Hoeller, F.~Farshidian, M.~Hutter, and A.~Garg, ``Articulated object interaction in unknown scenes with whole-body mobile manipulation,'' in \emph{2022 IEEE/RSJ international conference on intelligent robots and systems (IROS)}.\hskip 1em plus 0.5em minus 0.4em\relax IEEE, 2022, pp. 1647--1654.

\bibitem{mildenhall2021nerf}
B.~Mildenhall, P.~P. Srinivasan, M.~Tancik, J.~T. Barron, R.~Ramamoorthi, and R.~Ng, ``Nerf: Representing scenes as neural radiance fields for view synthesis,'' \emph{Communications of the ACM}, vol.~65, no.~1, pp. 99--106, 2021.

\bibitem{qureshi2024splatsim}
M.~N. Qureshi, S.~Garg, F.~Yandun, D.~Held, G.~Kantor, and A.~Silwal, ``Splatsim: Zero-shot sim2real transfer of rgb manipulation policies using gaussian splatting,'' \emph{arXiv preprint arXiv:2409.10161}, 2024.

\bibitem{lou2024robo}
H.~Lou, Y.~Liu, Y.~Pan, Y.~Geng, J.~Chen, W.~Ma, C.~Li, L.~Wang, H.~Feng, L.~Shi \emph{et~al.}, ``Robo-gs: A physics consistent spatial-temporal model for robotic arm with hybrid representation,'' \emph{arXiv preprint arXiv:2408.14873}, 2024.

\bibitem{lu2024manigaussian}
G.~Lu, S.~Zhang, Z.~Wang, C.~Liu, J.~Lu, and Y.~Tang, ``Manigaussian: Dynamic gaussian splatting for multi-task robotic manipulation,'' in \emph{European Conference on Computer Vision}.\hskip 1em plus 0.5em minus 0.4em\relax Springer, 2024, pp. 349--366.

\bibitem{zheng2024gaussiangrasper}
Y.~Zheng, X.~Chen, Y.~Zheng, S.~Gu, R.~Yang, B.~Jin, P.~Li, C.~Zhong, Z.~Wang, L.~Liu \emph{et~al.}, ``Gaussiangrasper: 3d language gaussian splatting for open-vocabulary robotic grasping,'' \emph{IEEE Robotics and Automation Letters}, 2024.

\bibitem{liu2024differentiable}
R.~Liu, A.~Canberk, S.~Song, and C.~Vondrick, ``Differentiable robot rendering,'' \emph{arXiv preprint arXiv:2410.13851}, 2024.

\bibitem{li2024robogsim}
X.~Li, J.~Li, Z.~Zhang, R.~Zhang, F.~Jia, T.~Wang, H.~Fan, K.-K. Tseng, and R.~Wang, ``Robogsim: A real2sim2real robotic gaussian splatting simulator,'' \emph{arXiv preprint arXiv:2411.11839}, 2024.

\bibitem{corke2007simple}
P.~I. Corke, ``A simple and systematic approach to assigning denavit--hartenberg parameters,'' \emph{IEEE transactions on robotics}, vol.~23, no.~3, pp. 590--594, 2007.

\bibitem{chen2024internvl}
Z.~Chen, J.~Wu, W.~Wang, W.~Su, G.~Chen, S.~Xing, M.~Zhong, Q.~Zhang, X.~Zhu, L.~Lu \emph{et~al.}, ``Internvl: Scaling up vision foundation models and aligning for generic visual-linguistic tasks,'' in \emph{Proceedings of the IEEE/CVF conference on computer vision and pattern recognition}, 2024, pp. 24\,185--24\,198.

\bibitem{xiang2020sapien}
F.~Xiang, Y.~Qin, K.~Mo, Y.~Xia, H.~Zhu, F.~Liu, M.~Liu, H.~Jiang, Y.~Yuan, H.~Wang \emph{et~al.}, ``Sapien: A simulated part-based interactive environment,'' in \emph{Proceedings of the IEEE/CVF Conference on Computer Vision and Pattern Recognition}, 2020, pp. 11\,097--11\,107.

\bibitem{li2020category}
X.~Li, H.~Wang, L.~Yi, L.~J. Guibas, A.~L. Abbott, and S.~Song, ``Category-level articulated object pose estimation,'' in \emph{Proceedings of the IEEE/CVF conference on computer vision and pattern recognition}, 2020, pp. 3706--3715.

\bibitem{dankwa2019twin}
S.~Fujimoto, H.~Hoof, and D.~Meger, ``Addressing function approximation error in actor-critic methods,'' in \emph{International conference on machine learning}.\hskip 1em plus 0.5em minus 0.4em\relax PMLR, 2018, pp. 1587--1596.

\end{thebibliography}

\end{document}